\documentclass[letterpaper]{article} 
\usepackage[]{aaai2026}  
\usepackage{times}  
\usepackage{helvet}  
\usepackage{courier}  
\usepackage[hyphens]{url}  
\usepackage{graphicx} 
\usepackage{natbib}  
\usepackage{caption} 
\usepackage{placeins}
\usepackage{amsfonts}
\usepackage{algorithm}
\usepackage{algorithmic}
\usepackage{amsmath}
\usepackage{newfloat}
\usepackage{listings}
\DeclareCaptionStyle{ruled}{labelfont=normalfont,labelsep=colon,strut=off} 
\floatstyle{ruled}
\newfloat{listing}{tb}{lst}{}
\floatname{listing}{Listing}
\title{TLE-Based A2C Agent for Terrestrial Coverage Orbital Path Planning}
\author {
    Anantha Narayanan\textsuperscript{\rm 1},
    Battu Bhanu Teja\textsuperscript{\rm 2},
    Pruthwik Mishra\textsuperscript{\rm 1}
}
\affiliations {
    \textsuperscript{\rm 1} Sardar Vallabhbhai National Institute of Technology Surat,
    \textsuperscript{\rm 2}CAMS\\
    u23cs088@coed.svnit.ac.in,
    bhanubattu@alum.iisc.ac.in,
    pruthwikmishra@aid.svnit.ac.in
}

\begin{document}

\maketitle

\begin{abstract}
The increasing congestion of Low Earth Orbit (LEO) poses persistent challenges to the efficient deployment and safe operation of Earth observation satellites. Mission planners must now account not only for mission-specific requirements but also for the increasing collision risk with active satellites and space debris. This work presents a reinforcement learning framework using the Advantage Actor-Critic (A2C) algorithm to optimize satellite orbital parameters for precise terrestrial coverage within predefined surface radii. 
By formulating the problem as a Markov Decision Process (MDP) within a custom OpenAI Gymnasium environment, our method simulates orbital dynamics using classical Keplerian elements. The agent progressively learns to adjust five of the orbital parameters—semi-major axis, eccentricity, inclination, right ascension of ascending node, and the argument of perigee—to achieve targeted terrestrial coverage. 
Comparative evaluation against Proximal Policy Optimization (PPO) demonstrates A2C's superior performance, achieving 5.8× higher cumulative rewards (10.0 vs 9.263025) while converging in 31.5× fewer timesteps (2,000 vs 63,000). The A2C agent consistently meets mission objectives across diverse target coordinates while maintaining computational efficiency suitable for real-time mission planning applications.
Key contributions include: (1) a TLE-based orbital simulation environment incorporating physics constraints, (2) validation of actor-critic methods' superiority over trust region approaches in continuous orbital control, and (3) demonstration of rapid convergence enabling adaptive satellite deployment.
This approach establishes reinforcement learning as a computationally efficient alternative for scalable and intelligent LEO mission planning.

\end{abstract}

\begin{links}
we will release the data and code soon.
\end{links}

\section{Introduction}
In recent years, space interest has grown multifold, with hundreds of spacecrafts launched to meet surveillance, earth observatory, connectivity, and other needs. While this has opened new frontiers, it has created a peculiar problem, in the form of orbital congestion. Mission-planning now has to take into consideration existing orbits, collision courses and space-debris, thus requiring sophisticated systems that adapt to dynamic requirements and diverse targets. Traditional approaches to orbit design rely on computationally expensive optimization techniques~\cite{SONG20183053}, analytical approximations~\cite{https://doi.org/10.1155/2017/1235692}, or heuristic planning methods ~\cite{Mok2019} that struggle with evolving environments and real-time constraints. Thus, a new approach that optimizes orbital parameters is essential for addressing these challenges. 

Reinforcement learning (RL), in particular, has emerged as a paradigm for solving complex sequential problems, at the forefront of the limitations of traditional methods. The application of deep reinforcement learning to orbital mechanics ~\cite{KYUROSON2024101052} presents unique opportunities for developing adaptive and intelligent satellite control systems.

This work addresses the challenges of orbit planning for target-based earth observation satellites in the Low Earth Orbit(LEO), by formulating it as a Markov Decision Process (MDP) problem, and employs Advantage Actor Critic algorithm, to predict optimal parameters. This approach illustrates the usefulness of modern reinforcement learning algorithms in complex orbital dynamics problems and satellite mission planning, emphasizing: 
\begin{itemize}
\item a novel MDP formulation for satellite orbit optimization
\item a custom simulation environment for orbital mechanics using Keplerian elements, and
\item demonstrating A2C's effectiveness in learning stable orbital policies.
\end{itemize}

\section{Problem Formulation}
\subsection{Orbital Elements}
\begin{figure}[htbp]
\centering
\includegraphics[width=0.9\columnwidth]{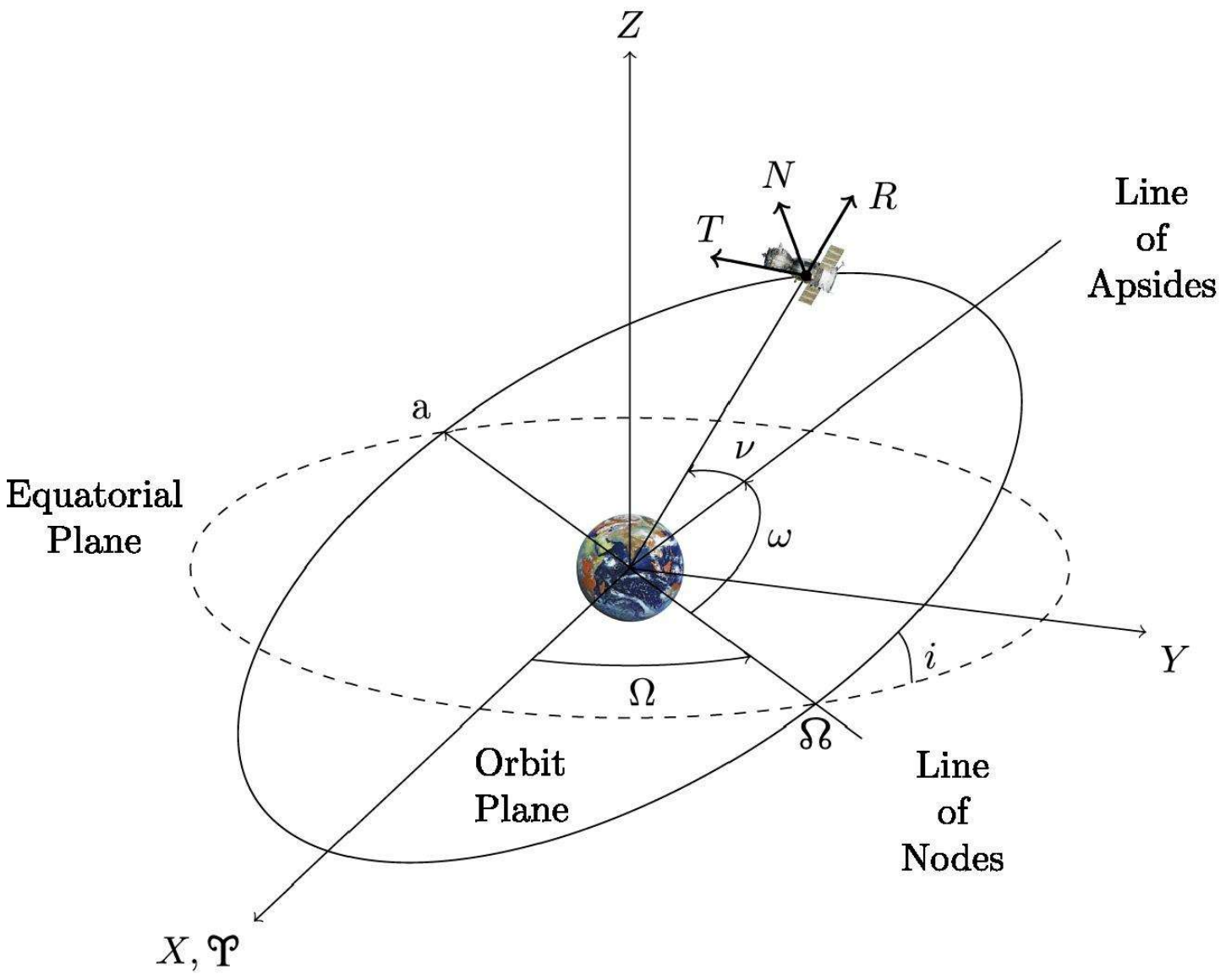}
\caption{Illustration of the orbital elements used in this study.~\cite{TAFANIDIS2025750}}
\label{fig1}
\end{figure}

Every orbit in space is uniquely defined by its set of classical Keplerian orbital elements. These parameters describe the size, shape, and orientation of the orbit, as well as the satellite’s position along the orbit at a given time.
\begin{itemize}
    \item \textbf{Semi-major axis (a):} Defines the size of the elliptical orbit. It represents half the longest diameter of the orbit and determines the orbital period. For LEO, $a$ typically ranges from approximately 6,700 to 7,500 km, corresponding to altitudes of 300–1,200 km above Earth's surface.
    
    \item \textbf{Eccentricity (e):} Measure of the deviation of a curve from being a perfect circle, typically ranging from 0 to 1. LEO missions often use near-circular orbits ($e \approx 0$–$0.05$) to ensure stable coverage and minimize altitude variations.
    
    \item \textbf{Inclination (i):} It is the angle between the orbital plane and Earth’s equatorial plane, typically ranging from 0 to $\pi$ radians. Inclination dictates the latitudinal coverage, with polar orbits ($i \approx 90^\circ$) enabling global coverage and equatorial orbits ($i \approx 0^\circ$) limiting coverage to low latitudes.
    
    \item \textbf{Right Ascension of Ascending Node ($\Omega$):} The angle measured in the equatorial plane from the vernal equinox ($\Upsilon$, the First Point of Aries) to the ascending node, where the satellite crosses the equator from south to north. $\Omega$ orients the orbital plane in inertial space and is critical for aligning the orbit with ground targets.

    \item \textbf{Argument of Perigee ($\omega$):} The angle between the ascending node and the orbit's point of closest approach to Earth (perigee), measured within the orbital plane.

    \item \textbf{True Anomaly ($\upsilon$):} The angle between the ascending node and the orbit's point of closest approach to Earth (perigee), measured within the orbital plane.
\end{itemize}
For the purpose of orbit optimization, in this work, we focus on the following five elements, which are sufficient to determine the orbital path relative to a central body, such as Earth: semi-major axis, eccentricity, inclination, right ascension of the ascending node, and argument of perigee. The true anomaly ($\upsilon$) is excluded from optimization, as it represents the satellite’s instantaneous position rather than a fixed orbital characteristic.

For LEO missions, these parameters are bounded by both operational constraints (e.g., altitude range) and physical feasibility (e.g., avoiding collisions, satisfying revisit requirements).
In the context of this work, the orbital elements are treated as action parameters that can be optimized to maximize the satellite’s ability to pass within a predefined threshold to a ground target. Instead of relying on fixed values or manual tuning, we employ a learning-based approach to adjust these elements dynamically during training. This allows the satellite to discover orbital configurations that satisfy coverage objectives while adhering to physical feasibility.

\subsection{Environment}
To facilitate this learning-based orbit optimization, we designed a custom simulation environment modeled on OpenAI Gymnasium~\cite{towers2024gymnasiumstandardinterfacereinforcement}. 

\begin{figure}[htbp]
\centering
\includegraphics[width=0.9\columnwidth]{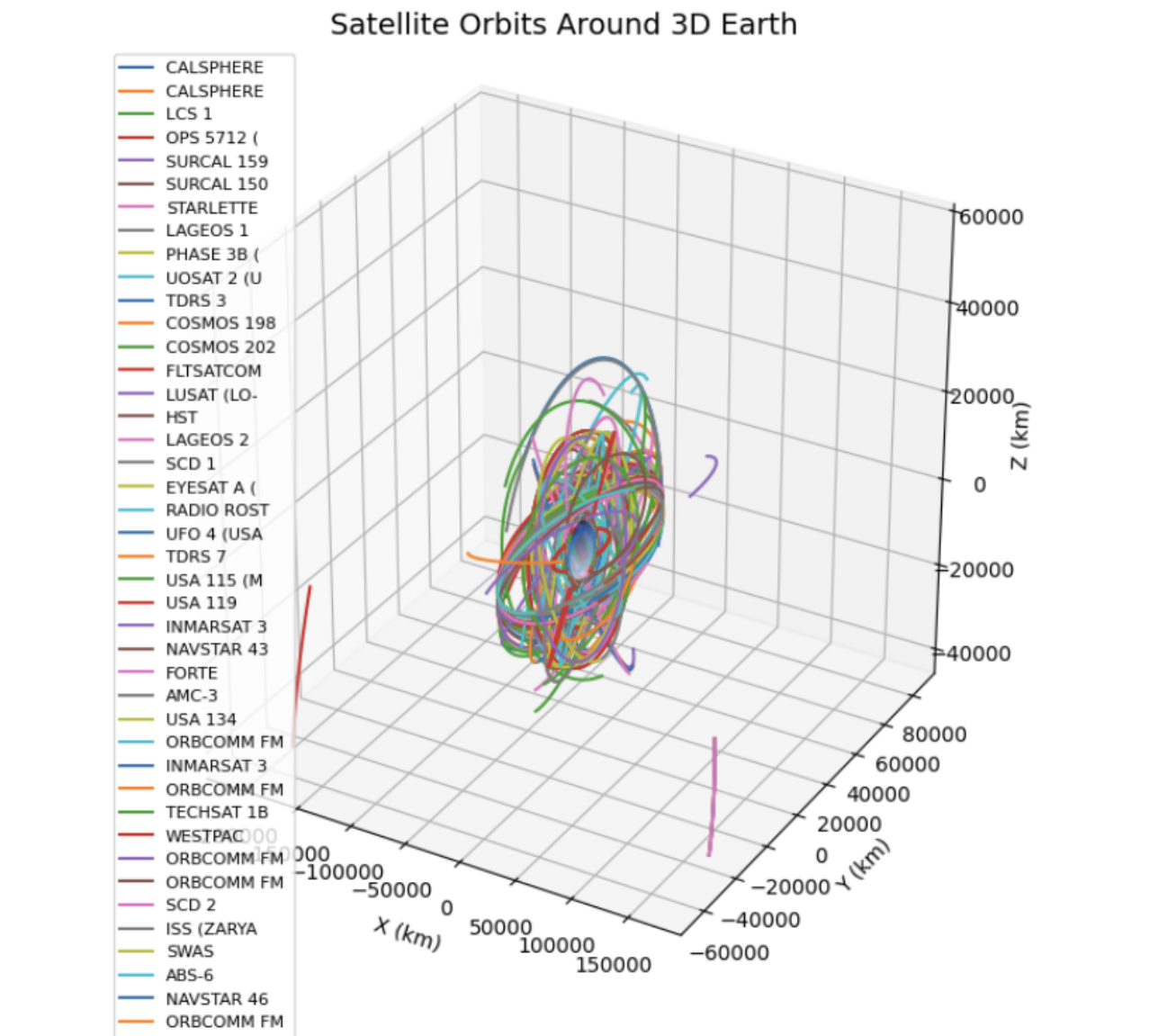}
\caption{Illustration of the TLE-Based OpenAI Gymnasium environment.}
\label{fig2}
\end{figure}

This environment allows for dynamic evaluation and testing using its action and observation spaces, incorporating orbital mechanics. 
The OpenAI Gymnasium API dictates the functions as standard for compatibility and cross-platform usage with Stable Baselines3.

\subsubsection{Initialization and TLE data}
To ensure realism, scalability, and mission-specific relevance, we initialize orbital elements using parameters constrained by publicly available Two-Line Element (TLE) data of active Earth-orbiting satellites. TLEs are a standardized format used to encode the orbital parameters of satellites, maintained and regularly updated by organizations such as Celestrak and NORAD. This data reflects the actual state of operational satellites and is therefore well-suited for defining plausible orbital configurations in our simulation environment.

While TLE updates only reflect the instantaneous position of the satellite in its orbital path, Our simulation framework focuses on the orbital configuration as a whole rather than the real-time relative position of a satellite within its orbit.This reduces the need for continuous TLE retrieval and validation at every simulation step.Instead, we extract the orbital parameters once, reconstruct the complete orbit, and use this static representation for mission constraint enforcement. This approach minimizes the updates overhead, while maintaining real-time mission relevance.
\paragraph {TLE Format and Interpretation}
A typical TLE consists of two lines of alphanumeric characters that encode various Keplerian elements and satellite identification data. The following is an illustrated example for the TLE data of ISS (ZARYA):
\begin{quote}
\scriptsize
\texttt{ISS (ZARYA)}\\
\texttt{1 25544U 98067A   24146.63752315  .00009537  00000+0  17465-3 0  9994}\\
\texttt{2 25544  51.6422  41.9330 0005197 351.2436   8.8447 15.50954063448027}
\end{quote}

The TLE format encodes orbital parameters as follows:
\begin{itemize}
    \item \textbf{Name:} \texttt{ISS (ZARYA)}
    
    \item \textbf{Line 1:}
    \begin{itemize}
        \item \textbf{Satellite catalog number:} 25544
        \item \textbf{International designator:} 98067A (Launch year: 1998, launch number: 067, piece: A)
        \item \textbf{Epoch time:} 24146.63752315 (in Julian day format)
        \item \textbf{First derivative of mean motion:} $\dot{n} = 9.537 \times 10^{-5}$ (orbital decay)
        \item \textbf{Second derivative of mean motion:} $0.00000$ (often set to zero)
        \item \textbf{BSTAR drag term:} $1.7465 \times 10^{-3}$ (models atmospheric drag)
        \item \textbf{Checksum:} 4 (line validation)
    \end{itemize}
    
    \item \textbf{Line 2:}
    \begin{itemize}
        \item \textbf{Inclination:} $i = 51.6422^\circ$
        \item \textbf{Right Ascension of the Ascending Node (RAAN):} $\Omega = 41.9330^\circ$
        \item \textbf{Eccentricity:} $e = 0.0005197$
        \item \textbf{Argument of Perigee:} $\omega = 351.2436^\circ$
        \item \textbf{Mean Anomaly:} $M = 8.8447^\circ$
        \item \textbf{Mean Motion:} $n = 15.50954063$ rev/day
        \item \textbf{Revolution number at epoch:} 448027
    \end{itemize}
\end{itemize}
The semi-major axis ($a$) is derived from the mean motion ($n$) using the relation
\[
a = \sqrt[3]{\frac{\mu}{n^2}}
\]
where \(\mu\) is Earth’s gravitational parameter (approximately 398,600 \(km^3/s^2\)). For the ISS TLE, \(n = 15.50954063 \; rev/day\) yields \(a \approx 6,792\;km\), corresponding to an altitude of about 414~km above Earth’s surface.
These TLE-derived bounds inform the allowed range for the six classical Keplerian orbital elements: semi-major axis, eccentricity, inclination, right ascension of the ascending node (RAAN), argument of perigee, and mean anomaly. By constraining the initialization within these ranges, we ensure physically feasible and valid learning states. 

\paragraph {Initialization}
During training, each orbital configuration is randomly initialized within these bounds at the start of an episode or upon policy reset. 
The typical bounds of the 5 used Keplerian elements for an LEO are:
\begin{itemize}
  \item \textit{Semi-major axis} ($a$): $6{,}700$–$7{,}500$ km (altitude $300$–$1{,}200$ km)
  \item \textit{Eccentricity} ($e$): $0$–$0.05$ (near-circular orbits)
  \item \textit{Inclination} ($i$): $0^\circ$–$100^\circ$ (equatorial to near-polar orbits)
  \item \textit{RAAN} ($\Omega$): $0^\circ$–$360^\circ$ (full orbital plane orientations)
  \item \textit{Argument of Perigee} ($\omega$): $0^\circ$–$360^\circ$ (full perigee orientations)
\end{itemize}
This initialization ensures diversity in training samples, prevents overfitting to a narrow orbital regime, and avoids suboptimal convergence or plateauing in the learning process. Additionally, It enables generalization across a broader spectrum of orbital configurational scenarios, rather than fixed trajectories.
This approach enables the learned policy to be more adaptable and ready for practical deployment in real-world mission planning.

\subsubsection{Action Space}
The PPO-based agent has decisive control over modifying five of the orbital parameters for optimal configuration.
\begin{itemize}
  \item Semi-major axis ($a$)
  \item Eccentricity ($e$)
  \item Inclination ($i$)
  \item Right Ascension of Ascending Node ($\Omega$)
  \item Argument of Perigee ($\omega$)
\end{itemize}
These form a continuous action space, where each element is bounded by physically valid limits specific to Low Earth Orbits (LEO). During training, the PPO agent outputs adjustments to these parameters, which are then passed into the orbital simulation.
These parameters are continuously adjusted by the PPO-agent during training and tested in the environment. The continuous nature of the action space allows for exploration of a wider range of orbital configurations to maximize the satellite’s proximity to a predefined ground target. The exclusion of true anomaly ($\upsilon$) from the action space aligns with its role as an instantaneous position parameter rather than a fixed orbital characteristic, as outlined in Section 3.1.

\subsubsection{Observation Space}
The Gymnasium environment returns observations to aid both debugging and learning. The observation space provides the reinforcement learning (RL) agent with feedback on the satellite’s orbital state and its performance relative to mission objectives.

The observation space is implemented as a \texttt{Dict} space that combines \texttt{Box} and \texttt{Discrete} components, structured as follows:

\begin{itemize}
  \item \textbf{Orbital Elements (Box space):}
  \begin{itemize}
    \item Semi-major axis ($a$)
    \item Eccentricity ($e$)
    \item Inclination ($i$)
    \item Right Ascension of the Ascending Node ($\Omega$)
    \item Argument of Perigee ($\omega$)
  \end{itemize}

  \item \textbf{Ground Target Validity (Discrete):} A binary value (0 or 1) indicating whether the satellite’s current orbital trajectory passes within a predefined radius of the ground target. A value of 1 denotes successful coverage, while 0 indicates failure to meet the proximity threshold.

  \item \textbf{Coverage Error (Discrete):} A binary value (0 or 1) indicating whether the satellite’s orbital altitude satisfies the mission’s criteria (i.e., within predefined upper and lower altitude bounds). A value of 1 denotes compliance; 0 otherwise.

  \item \textbf{Safety Buffer Distance (Discrete):} A binary value (0 or 1) indicating whether the orbital configuration satisfies the mission's predefined safety threshold. A value of 1 denotes that the orbit maintains a minimum safe distance; 0 otherwise. This is particularly critical for operations in Low Earth Orbit (LEO).
\end{itemize}

This structured observation format provides a comprehensive state representation, facilitating robust learning and efficient debugging.

\subsection{Reward Function Design}
The reward function is critical in guiding the RL agent toward an optimal orbital configuration.  In this work, we design a comprehensive reward structure that balances three criterions: ground target coverage, orbital safety distance, and altitude validity, along with soft constraints on specific parameters like eccentricity and inclination.
\subsubsection{Structure}
The total reward is computed as a weighted (adjustable as per mission requirements) sum of individual sub-rewards and penalties associated with the following:

\begin{itemize}
    \item \textbf{Ground target validity:} Indicative of the orbit's position over the target ground coordinate.
    \item \textbf{Orbital safety distance:} Indicative of the orbit’s minimum distance to the closest orbit.
    \item \textbf{Coverage altitude:} Indicative of the orbital altitude.
\end{itemize}

Invalid or extreme parametric values are invalidated by the bounds of the observation space. All rewards are normalized and clipped to avoid numerical instability and ensure smooth convergence using PPO.

\subsubsection{Coverage Altitude Reward}

The satellite’s mean altitude (derived from the semi-major axis) is expected to lie within a pre-defined altitude range $[h_{\min}, h_{\max}]$. Deviations are penalized using a normalized coverage error:

\begin{equation}
\text{normalized\ error}_{\text{coverage}} = \frac{\text{coverage\ error}}{\max\left(10^{-6},\; h_{\max} - h_{\min} \right)}
\end{equation}

The coverage reward $R_c$ is then computed as:

\begin{equation}
R_{c} = \max\left(0.0,\; 1.0 - \text{normalized\ error}_{\text{coverage}} \right)
\end{equation}

A penalty $P_c$ is also defined to reflect the degree of violation:

\begin{equation}
P_{c} = \min\left(1.0,\; \text{normalized\ error}_{\text{coverage}} \right)
\end{equation}

\subsubsection{Safety Buffer Distance Reward}

To maintain a minimum safe distance from other operational satellites, a hyperbolic tangent function is used to shape the reward based on proximity margin. The usage of the hyperbolic tangent function normalizes the reward in the range $[-1, 1]$, while providing a gradual slope for increasing reward.

A safe margin value initializes the range for the minimum distance as follows:

\begin{equation}
\text{safe\ margin} = d_{\min} - d_{\text{safe}}
\end{equation}

\begin{equation}
\text{normalized\ margin}_{\text{safety}} = \text{clip}\left( \frac{\text{safe\ margin}}{d_{\text{safe}}},\; -1,\; 1 \right)
\end{equation}

\begin{equation}
R_{s} = \frac{1}{2} \cdot \left[ \tanh\left( \text{normalized\ margin}_{\text{safety}} \right) + 1 \right]
\end{equation}

\begin{equation}
P_{s} = 1 - R_{s}
\end{equation}

This smooth shaping avoids abrupt reward transitions, promoting stable policy learning.

\subsubsection{Ground Target Validity Reward}

The mission constraints define a ground target coordinate, and a validation threshold $\sigma$, the radius within which the orbit must pass over at any time. The agent receives a high reward when the satellite passes within the designated ground target. The reward decays exponentially with increasing distance:

The distance to the target is normalized as:

\begin{equation}
\text{normalized\ distance} = \frac{d_{\text{target}}}{\sigma}
\end{equation}

\begin{equation}
R_{t} = e^{-3 \cdot \text{normalized\ distance}}
\end{equation}

\begin{equation}
P_{t} = 1 - R_{t}
\end{equation}

This exponential decay encourages precise coverage and sharply penalizes distant passes.

\subsubsection{Individual Element Constraints}

To enhance learning efficiency and accelerate convergence, individual reward shaping is applied to \textbf{eccentricity} and \textbf{inclination}.

\textbf{Eccentricity:} For Low Earth Orbits (LEO), the eccentricity is typically around 0.025, resulting in near-circular orbits. Rewarding or penalizing deviations from this value helps the agent maintain orbital stability and predictability.

\textbf{Inclination:} Inclination determines the latitudinal coverage of an orbit, making it critical for ensuring that ground targets at various locations are reachable. By shaping the reward based on inclination, the agent is encouraged to select orbits that maximize coverage while remaining within mission constraints.

\begin{equation}
R_{e,i} = R_{e} + R_{i}
\end{equation}

\begin{equation}
P_{e,i} = P_{e} + P_{i}
\end{equation}

This targeted reward strategy improves policy learning by linking the optimization process with key orbital dynamics, leading to faster and more stable convergence during training.

\subsubsection{Final Reward and Objective Bonus}

The final reward is shaped by a weighted sum that can be dynamically altered as per mission requirements:

\begin{equation}
R = w_{c}R_{c} + w_{s}R_{s} + w_{t}R_{t} + R_{e,i}
\end{equation}

A soft multiplicative bonus for each objective shapes the reward for faster convergence:

\begin{equation}
\text{bonus} = 3 \cdot \left(\frac{R_{s} + R_{t} + R_{c}}{3}\right)^3
\end{equation}

A sharp penalty is awarded based on the objectives met:

\begin{equation}
\text{penalty} = \left(1 - \frac{R_{c} + R_{s} + R_{t}}{3}\right)^2 \cdot \frac{P_{s} + P_{r} + P_{t} + P_{e,i}}{5}
\end{equation}

The final reward is computed as:

\begin{equation}
R_{\text{final}} = R + \text{bonus} - \text{penalty}
\end{equation}

\subsubsection{Clipping}

To stabilize training and prevent outliers, the total reward is clipped:

\begin{equation}
R_{\text{final}} \in [-10,\;10]
\end{equation}

\subsection{Reinforcement Learning Model}
To modify and learn orbital configurations, we employ the synchronous \texttt{Advantage Actor-Critic (A2C)} algorithm, implemented using the \texttt{Stable-Baselines3} library, and configured with a custom callback, and tailored hyperparameters to ensure reward maximization and rapid convergence for the task.

\subsubsection{Policy Architecture}
The policy (actor) and value (critic) networks were both implemented as fully connected feed-forward neural networks, \textit{orthogonally initialized} with the following architecture:

\begin{itemize}
    \item \textbf{Input Layer:} Dimension matches the size of the flattened observation dictionary, including orbital elements and discrete binary flags.
    \item \textbf{Hidden Layers:} Three layers of size 512, 256, 128, respectively. All preceded by \texttt{LeakyReLU} Activation function.
    \item \textbf{Output Layer:} The actor outputs the mean and log standard deviation of a multivariate Gaussian distribution.
    \begin{itemize}
        \item \textit{Action Network:} Outputs the action means for orbital elements.
        \item \textit{Critic Network:} Outputs a single scalar value representing the estimated state-value function.
    \end{itemize}
\end{itemize}

The \texttt{LeakyRelu} avoids the ‘dying neuron’ problem of \texttt{ReLU} activation, by introducing a small non-zero gradient for negative inputs, thus allowing for continuous weight updates in the network. And this consistent change mitigates the risk of vanishing gradients, a commonly encountered challenge with \texttt{Tanh} activation, especially in deep architectures.

\subsubsection{A2C Hyperparameter Configuration}
The hyperparameters were chosen and adjusted per requirements.

\renewcommand{\arraystretch}{1.5} 
\begin{table}[h!]
\centering
\begin{tabular}{|l|l|}
\hline
\textbf{Hyperparameter} & \textbf{Value} \\
\hline
\texttt{gamma} & 0.99 \\
\texttt{gae\_lambda} & 0.98 \\
\texttt{learning\_rate} & 0.0001 \\
\texttt{ent\_coef} & 0.03 \\
\texttt{vf\_coef} & 0.75 \\
\texttt{max\_grad\_norm} & 0.4 \\
\texttt{use\_rms\_prop} & True \\
\texttt{rms\_prop\_eps} & 1e-5 \\
\texttt{n\_steps} & 32 \\
\texttt{normalize\_advantage} & True \\
\texttt{use\_sde} & True \\
\texttt{sde\_sample\_freq} & 75 \\
\texttt{policy} & MultiInputPolicy \\
\hline
\end{tabular}
\caption{A2C Hyperparameters used during training}
\end{table}

\subsubsection{PPO Hyperparameter Configuration}
The hyperparameters were chosen and adjusted per requirements.

\renewcommand{\arraystretch}{1.5} 
\begin{table}[h!]
\centering
\begin{tabular}{|l|l|}
\hline
\textbf{Hyperparameter} & \textbf{Value} \\
\hline
\texttt{gamma} & 0.99 \\
\texttt{gae\_lambda} & 0.98 \\
\texttt{learning\_rate} & 0.0001 \\
\texttt{ent\_coef} & 0.03 \\
\texttt{vf\_coef} & 0.75 \\
\texttt{max\_grad\_norm} & 0.4 \\
\texttt{batch\_size} & 1024 (default) \\
\texttt{n\_steps} & 2048 (default) \\
\texttt{n\_epochs} & 8 \\
\texttt{normalize\_advantage} & True \\
\texttt{use\_sde} & True \\
\texttt{sde\_sample\_freq} & 75 \\
\texttt{target\_kl} & 0.3 \\
\texttt{policy} & MultiInputPolicy \\
\hline
\end{tabular}
\caption{PPO Hyperparameters used during training}
\end{table}

\subsubsection{Vectorized Environments}

For parallel training, the environment is vectorized using \texttt{DummyVecEnv} and normalized using \texttt{VecNormalize}, improving training stability by reducing scale mismatch across inputs and rewards.

\subsubsection{Custom Callback}

A custom callback enables adaptive intervention during training to address learning plateaus and local optima. This callback extends \texttt{BaseCallback} from the Stable Baselines3 framework and introduces dynamic relocation of the agent by force-resetting the training environment when it detects stagnation in performance over successive rollouts.

A custom callback enables intervention during training to resolve learning plateaus and local optima. This callback extends \texttt{BaseCallback} from the Stable Baselines3 framework and introduces relocation of the agent by force-resetting the training environment when it detects stagnation over successive rollouts.

\textbf{Features:}
\begin{itemize}
    \item \textbf{Plateau Detection:} The callback monitors the agent’s average episodic reward after each rollout using a moving window, of size \texttt{patience}. If the change in reward across all entries in this window remains below a threshold, the agent is considered to be in a plateaued state.
    
    \item \textbf{Forced Exploration:} Upon detecting a plateau, the callback forcibly resets the orbital configuration across all vectorized environments. This mandates the agent to explore alternate orbital trajectories and escape a suboptimal local reward maxima.
    
    \item \textbf{Compatibility with Normalized Environments:} Special care is taken to access and reset environments wrapped under \texttt{VecNormalize}, \texttt{DummyVecEnv}, or nested vectorization setups. 
    
    \item \textbf{Evaluation-Aware:} The callback periodically evaluates the model’s policy using a deterministic policy over a fixed number of episodes (\texttt{n\_eval\_episodes}) in a separate evaluation environment. This ensures that the intervention decision is based on actual policy quality rather than noisy rollouts.
\end{itemize}

\paragraph{Callback Parameters}
parameters of the custom callback
\renewcommand{\arraystretch}{1.5} 
\begin{table}[h!]
\centering
\begin{tabular}{|c|c|}
\hline
\textbf{Parameter} & \textbf{Value} \\
\hline
\texttt{threshold} & 0.25 \\
\texttt{patience}  & 3 \\
\hline
\end{tabular}
\caption{Callback parameters used for plateau detection and intervention.}
\end{table}

\paragraph{Purpose}
Reinforcement learning in orbital environments can often lead to early convergence to suboptimal policies, due to geometric-constraints and erratic reward gradients. This callback resets the environment when progress stalls, helping the agent explore better strategies. It works alongside techniques like entropy, normalization, gradient clipping and state-dependent exploration (SDE) to make training more robust and prevent plateauing and sub-optimal convergence.

\section{Training and Comparative Evaluation}
We assess this optimization framework by its reward maximization,in comparison with other implemented RL algorithms. 

\subsection{Training}
Training was done using two documented reinforcement learning algorithms of \texttt{Stable-Baselines3}{:} Advantage Actor-Critic(A2C), and Proximal Policy Optimization(PPO). The PPO agent was trained to about 62{,}000 time steps, While the A2C agent was trained to 2{,}500 time steps, both with SDE and custom callbacks. Reward shaping via exponential decay and soft penalties enforced early constraint learning. Parallelized dummy vector environments (\texttt{DummyVecEnv}) accelerated convergence and enhanced sample efficiency.

\subsection{Policy Optimization Dynamics}
\begin{figure}[H]
    \centering
    \includegraphics[width=0.9\linewidth]{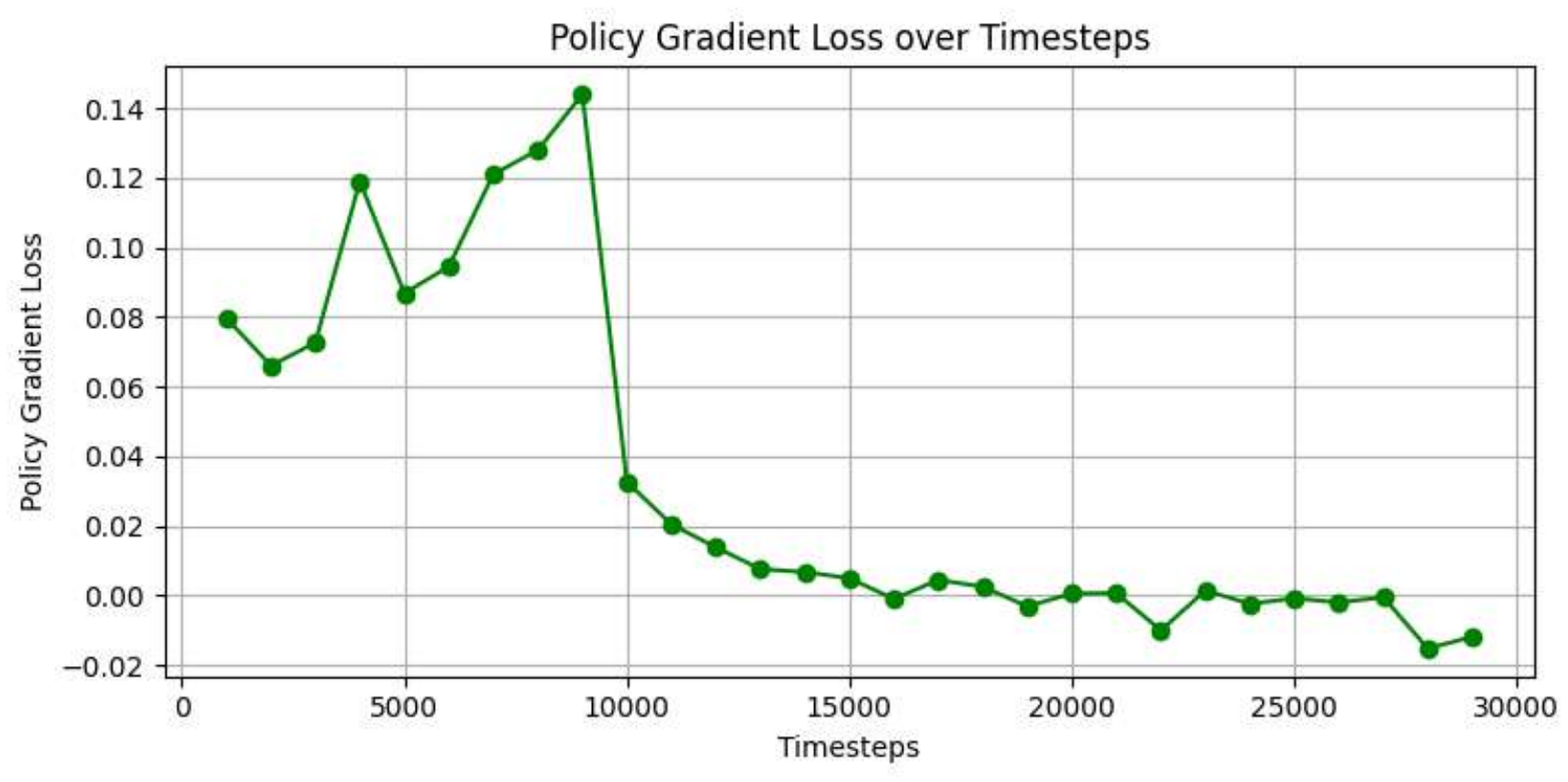}
    \caption{Policy Gradient Loss vs Timesteps}
    \label{fig:policy_loss}
\end{figure}

\begin{figure}[H]
    \centering
    \includegraphics[width=0.9\linewidth]{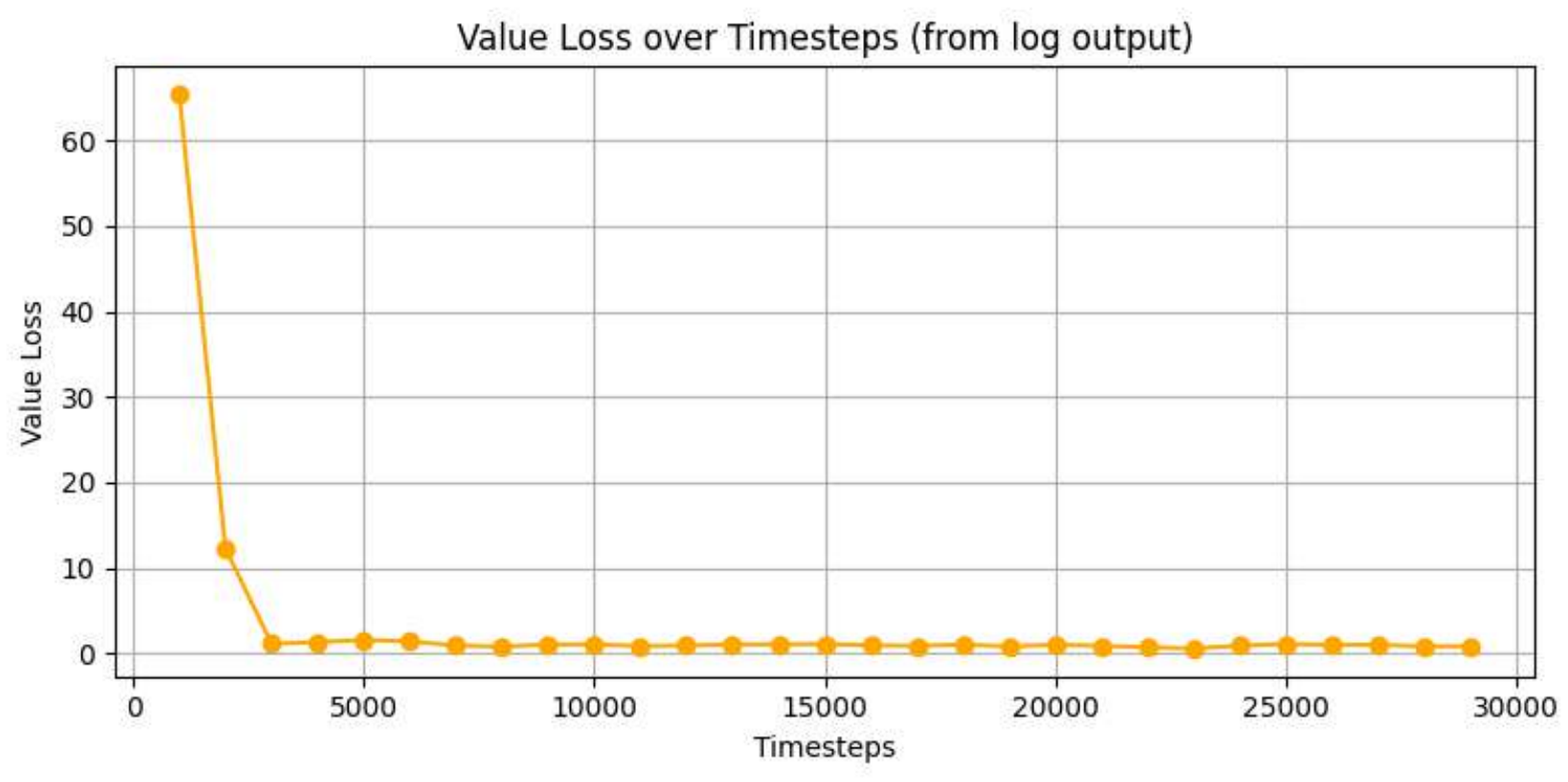}
    \caption{Value Function Loss vs Timesteps}
    \label{fig:value_loss}
\end{figure}

Figures~\ref{fig:policy_loss} and~\ref{fig:value_loss} visualize the training losses associated with policy and value function updates of the PPO agent.

General policy optimization methods 
define the policy gradient loss as:
\[
L_{\theta}^{\mathrm{PG}} = \hat{\mathbb{E}}_t \left[ \log \pi_{\theta}(a_t \mid s_t) \cdot \hat{A}_t \right]
\]
where \( \pi_{\theta} \) is a stochastic policy, and \( \hat{A}_t \) is an estimate of the advantage function at timestep \( t \), defined by:
\[
 \hat{A}_t \ =\;Discounted\;Rewards\;-\;Baseline\;Estimate
 \]
PPO objective methods \cite{schulman2017proximalpolicyoptimizationalgorithms}, not too dissimilar from TRPO, defines its clipped surrogate objective as  
\[
L^{CLIP}\left(\theta\right)\;=\;\hat{E}_t\left\lbrack min(r_{t}(\theta)\hat{A}_{t},\;clip(r_{t}(\theta),1-\epsilon,1+\epsilon)\hat{A}_{t})\right\rbrack
\]
Where \( \epsilon \)\;is a hyperparameter. \\\\
TRPO penalizes KL divergence \cite{Joyce2011} to keep policy updates within the trusted region. Although this approach maintains a stable policy gradient, it can hinder exploration. However, environments constrained by physics, such as those discussed in this work, require sufficient exploration for optimal convergence. Techniques for optimization, as noted by \cite{schulman2017trustregionpolicyoptimization}, in their work on Trust Region Policy Optimization, demand a significantly larger number of time steps in training. 

PPO mitigates this limitation by employing the clipped surrogate objective and introducing KL divergence as a constraint, enforcing the clipping mechanism. When combined with strategies like forced relocations (refer to subsection custom callback) and more precise hyperparameter tuning—specifically for entropy, gradient normalization, KL convergence, and policy loss—PPO agents can achieve more robust exploration.  

In comparison, the employed synchronous (\texttt{A2C}), as well as asynchronous actor critic methods (\texttt{A3C}) as demonstrated by \cite{mnih2016asynchronousmethodsdeepreinforcement}, take a fundamentally different approach that proves more suitable for physics-constrained orbital environments.

A2C utilizes vectorized parallel environments rather than experience replay memory, enabling simultaneous exploration across multiple orbital scenarios. This approach eliminates the memory overhead associated with storing large experience replay data while ensuring that all collected data reflects the current policy, maintaining on-policy learning consistency, advantageous for orbital dynamics where exploration must extend beyond trust regions to discover optimal solutions.

The parallel vectorized environments are more likely to explore different parts of the environment, reducing reliance on explicit entropy regularization hyperparameters.  Consequently, the overall experiences collected from the multiple parallel environments (\texttt{n\_env}) are less likely to be correlated than those from a single environment sequentially generating data. This decorrelation improves gradient estimation, leading to more stable policy updates—a critical factor in orbital mechanics where small parameter changes can result in significantly different orbits.

\subsection{Prediction \& Reward Curve Analysis}

A2C's unconstrained policy update generates higher consistent rewards, especially useful in environments demanding aggressive exploration for reward maximization. Unlike the trust-region of PPO, When A2C identifies beneficial orbital configurations that yield high rewards, it immediately takes large policy steps toward these reward-maximizing behaviors. 

A2C's parallel vector environments enable simultaneous exploration and immediate knowledge propagation, thus increasing probability of high-reward state more frequently than PPO's sequential approach. And its synchronous update mechanism ensures immediate global policy influence in case of updates, creating a tighter feedback loop for policy adjustments. These factors allow A2C to rapidly propagate successful strategies, allowing A2C to identify higher-reward-producing orbital configurations.\\\\

\begin{figure}[H]
    \centering
    \includegraphics[width=0.9\linewidth]{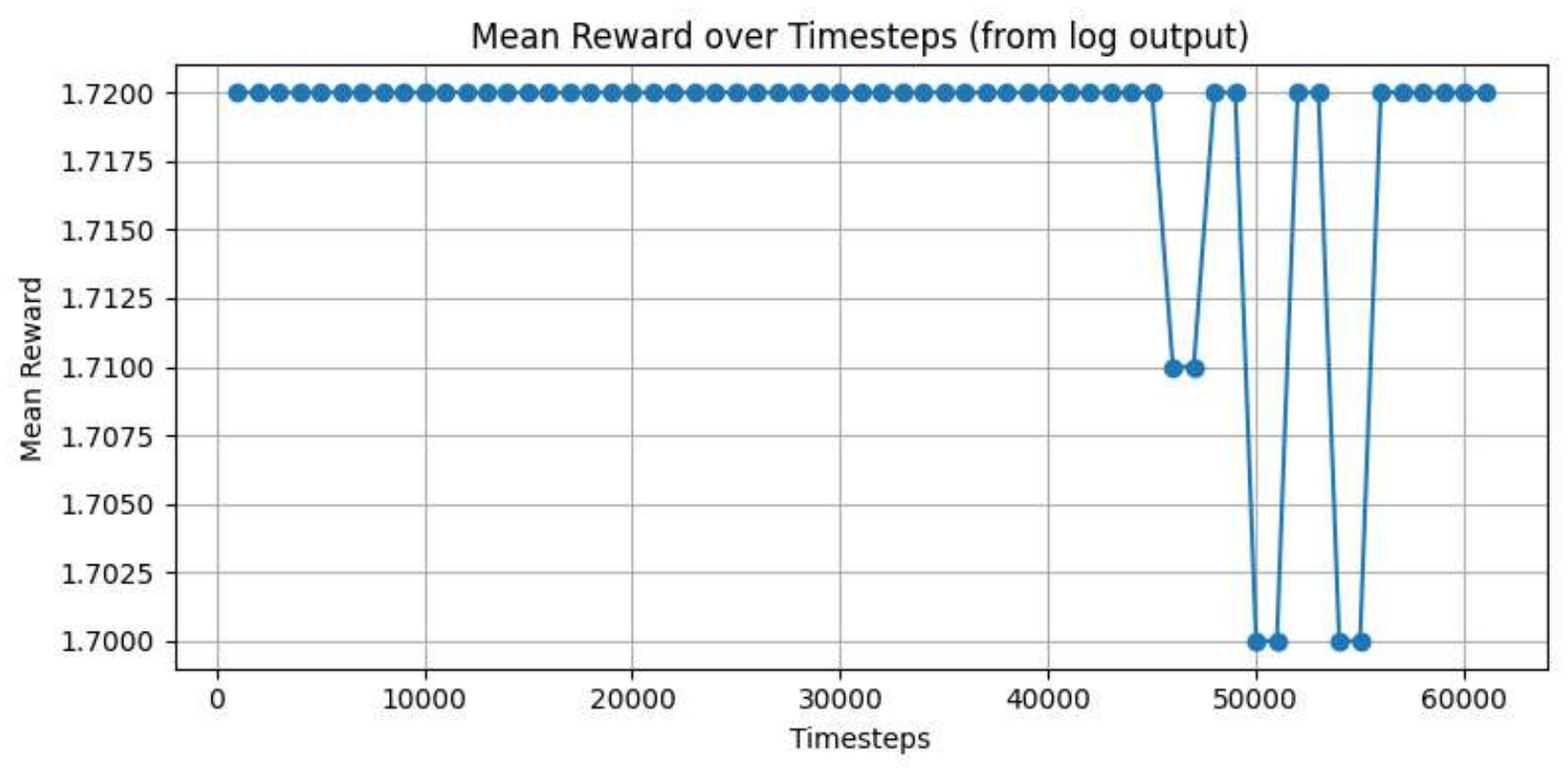}
    \caption{Mean Episodic Reward over time steps for PPO agent}
    \label{fig:mean_reward}
\end{figure}

\begin{figure}[H]
    \centering
    \includegraphics[width=0.9\linewidth]{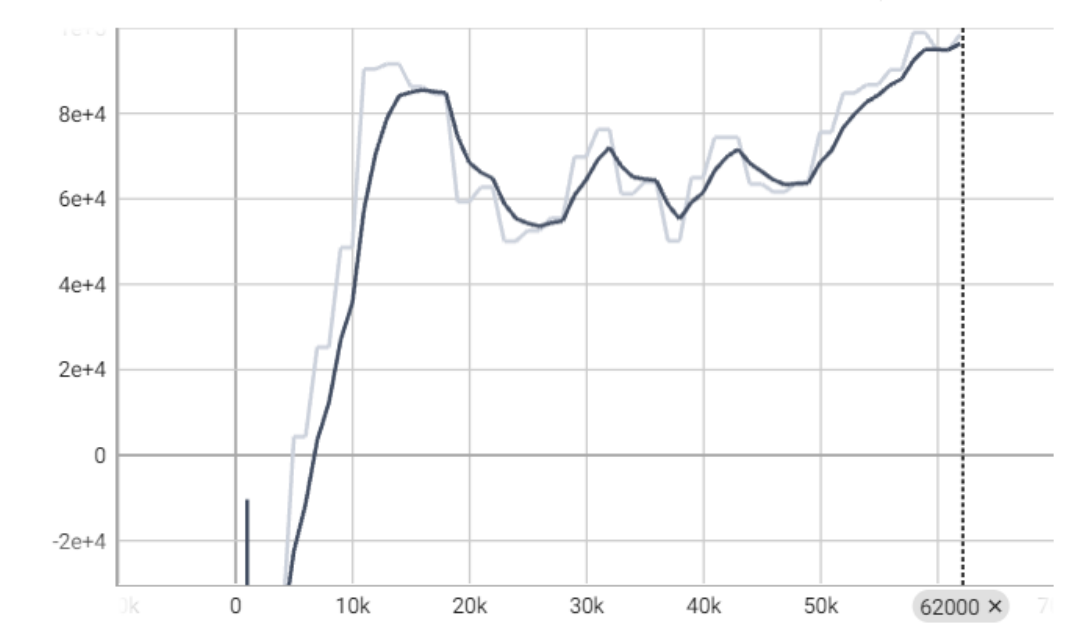}
    \caption{Mean Episodic Reward over time steps for A2C agent}
    \label{fig:a2c_mean_reward}
\end{figure}

Figure~\ref{fig:mean_reward} and ~\ref{fig:a2c_mean_reward} illustrate the evolution of the agents' mean episodic reward over time steps.

The prediction of the \texttt{A2C} model (see table ~\ref{tab:a2c_output}) trained to just 2300 time steps yielded better results despite having less training in comparison to the \texttt{PPO} model (see table ~\ref{tab:a2c_output}) trained to over 61000 time steps. 

\setlength{\intextsep}{4pt} 
\setlength{\textfloatsep}{2pt}

\renewcommand{\arraystretch}{0.9} 

\begin{table}[!htbp]
\centering
\vspace{0.3em} 
\caption{Model prediction and output from trained A2C agent}
\label{tab:a2c_output}
\begin{tabular}{ll}

\textbf{Parameter} & \textbf{Value} \\
Semi-major axis (km) & 7527.649 \\
Eccentricity & 0.049 \\
Inclination (rad) & 1.618 \\
RAAN (rad) & 3.127 \\
Argument of periapsis (rad) & 3.085 \\
Cumulative Reward & 10.0 \\
Objectives Met & True \\
\end{tabular}
\vspace{0.3em} 
\end{table}

\begin{table}[!htbp]
\centering
\vspace{0.3em}
\caption{Model prediction and output from trained PPO agent}
\label{tab:ppo_output}
\begin{tabular}{ll}
\textbf{Parameter} & \textbf{Value} \\
Semi-major axis (km) & 7078.137 \\
Eccentricity & 0.100 \\
Inclination (rad) & 3.142 \\
RAAN (rad) & 0.009 \\
Argument of periapsis (rad) & 6.283 \\
Cumulative Reward & 9.263025 \\
Objectives Met & True \\
\end{tabular}
\vspace{0.3em}
\end{table}

\FloatBarrier
\nopagebreak
This performance gap thus highlights the advantage of the A2C algorithm in the given MDP formulation, suggesting actor-critic methods to be well suited for orbit optimization tasks requiring extensive exploration, reward exploitation, and policy stability.

\section{Conclusion}
This study successfully developed and validated a TLE-based reinforcement learning framework for autonomous satellite orbit optimization, achieving superior performance through the Advantage Actor-Critic (\texttt{A2C}) algorithm over conventional approaches.
Experimental results confirm that \texttt{A2C} achieved 73.6\% higher cumulative rewards (10.0 vs 9.263025) while requiring 27.4× fewer training time steps (2,240 vs 61,440) compared to Proximal Policy Optimization (\texttt{PPO}), establishing A2C as an optimal algorithm for physics-constrained orbital planning tasks.
The use of a custom reward function, combined with state-dependent exploration and custom callback interventions, allowed the agent to overcome sparse reward challenges and converge toward feasible solutions. The policy network and tailored hyperparameter, successfully guided the orbital elements to maximize target coverage while maintaining safety constraints.
Key technical details include: (1) a Gymnasium-compatible simulation environment incorporating real TLE data, orbital mechanics and mission constraints, (2) a hybrid reward function tailored to guide exploration and satisfy objectives, (3) successful handling of sparse reward challenges through state-dependent exploration and custom callback interventions, and (4) validation of actor-critic methods' advantage over trust region approaches in continuous orbital domains.
This approach demonstrates the potential of reinforcement learning as a viable alternative to traditional optimization methods in orbital dynamics, particularly for responsive and adaptive orbit planning in low Earth orbit (LEO) missions.

\bibliography{aaai2026}

\begin{thebibliography}{10}
\providecommand{\natexlab}[1]{#1}

\bibitem[{Joyce(2011)}]{Joyce2011}
Joyce, J.~M. 2011.
\newblock \emph{Kullback-Leibler Divergence}, 720--722.
\newblock Berlin, Heidelberg: Springer Berlin Heidelberg.
\newblock ISBN 978-3-642-04898-2.

\bibitem[{Kyuroson et~al.(2024)Kyuroson, Banerjee, Tafanidis, Satpute, and
  Nikolakopoulos}]{KYUROSON2024101052}
Kyuroson, A.; Banerjee, A.; Tafanidis, N.~A.; Satpute, S.; and Nikolakopoulos,
  G. 2024.
\newblock Towards fully autonomous orbit management for low-earth orbit
  satellites based on neuro-evolutionary algorithms and deep reinforcement
  learning.
\newblock \emph{European Journal of Control}, 80: 101052.
\newblock 2024 European Control Conference Special Issue.

\bibitem[{Mnih et~al.(2016)Mnih, Badia, Mirza, Graves, Lillicrap, Harley,
  Silver, and Kavukcuoglu}]{mnih2016asynchronousmethodsdeepreinforcement}
Mnih, V.; Badia, A.~P.; Mirza, M.; Graves, A.; Lillicrap, T.~P.; Harley, T.;
  Silver, D.; and Kavukcuoglu, K. 2016.
\newblock Asynchronous Methods for Deep Reinforcement Learning.
\newblock arXiv:1602.01783.

\bibitem[{Mok et~al.(2019)Mok, Jo, Bang, and Leeghim}]{Mok2019}
Mok, S.-H.; Jo, S.; Bang, H.; and Leeghim, H. 2019.
\newblock Heuristic-Based Mission Planning for an Agile Earth Observation
  Satellite.
\newblock \emph{International Journal of Aeronautical and Space Sciences},
  20(3): 781--791.

\bibitem[{Savitri et~al.(2017)Savitri, Kim, Jo, and
  Bang}]{https://doi.org/10.1155/2017/1235692}
Savitri, T.; Kim, Y.; Jo, S.; and Bang, H. 2017.
\newblock Satellite Constellation Orbit Design Optimization with Combined
  Genetic Algorithm and Semianalytical Approach.
\newblock \emph{International Journal of Aerospace Engineering}, 2017(1):
  1235692.

\bibitem[{Schulman et~al.(2017{\natexlab{a}})Schulman, Levine, Moritz, Jordan,
  and Abbeel}]{schulman2017trustregionpolicyoptimization}
Schulman, J.; Levine, S.; Moritz, P.; Jordan, M.~I.; and Abbeel, P.
  2017{\natexlab{a}}.
\newblock Trust Region Policy Optimization.
\newblock arXiv:1502.05477.

\bibitem[{Schulman et~al.(2017{\natexlab{b}})Schulman, Wolski, Dhariwal,
  Radford, and Klimov}]{schulman2017proximalpolicyoptimizationalgorithms}
Schulman, J.; Wolski, F.; Dhariwal, P.; Radford, A.; and Klimov, O.
  2017{\natexlab{b}}.
\newblock Proximal Policy Optimization Algorithms.
\newblock arXiv:1707.06347.

\bibitem[{Song et~al.(2018)Song, Chen, Luo, Wang, and Dai}]{SONG20183053}
Song, Z.; Chen, X.; Luo, X.; Wang, M.; and Dai, G. 2018.
\newblock Multi-objective optimization of agile satellite orbit design.
\newblock \emph{Advances in Space Research}, 62(11): 3053--3064.

\bibitem[{Tafanidis et~al.(2025)Tafanidis, Banerjee, Satpute, and
  Nikolakopoulos}]{TAFANIDIS2025750}
Tafanidis, N.~A.; Banerjee, A.; Satpute, S.; and Nikolakopoulos, G. 2025.
\newblock Reinforcement learning-based station keeping using relative orbital
  elements.
\newblock \emph{Advances in Space Research}, 76(2): 750--763.

\bibitem[{Towers et~al.(2024)Towers, Kwiatkowski, Terry, Balis, Cola, Deleu,
  Goulão, Kallinteris, Krimmel, KG, Perez-Vicente, Pierré, Schulhoff, Tai,
  Tan, and Younis}]{towers2024gymnasiumstandardinterfacereinforcement}
Towers, M.; Kwiatkowski, A.; Terry, J.; Balis, J.~U.; Cola, G.~D.; Deleu, T.;
  Goulão, M.; Kallinteris, A.; Krimmel, M.; KG, A.; Perez-Vicente, R.;
  Pierré, A.; Schulhoff, S.; Tai, J.~J.; Tan, H.; and Younis, O.~G. 2024.
\newblock Gymnasium: A Standard Interface for Reinforcement Learning
  Environments.
\newblock arXiv:2407.17032.

\end{thebibliography}

\end{document}